\documentclass[conference, 10pt]{IEEEtran}
\IEEEoverridecommandlockouts
\usepackage{cite}
\usepackage{amsmath,amssymb,amsfonts}
\usepackage{algorithmic}
\usepackage{graphicx}
\usepackage{textcomp}
\usepackage{xcolor}
\usepackage{url}
\usepackage{placeins}
\usepackage{float}
\usepackage{tabularx,colortbl}
\usepackage{ifthen}
\usepackage{caption,subcaption}
\usepackage{amsmath}
\usepackage{booktabs}
\renewcommand{\thetable}{\Roman{table}}
\usepackage{cleveref}

\usepackage{enumitem}
\usepackage{stfloats}
\usepackage{wrapfig}
\usepackage{verbatim}
\usepackage{tabularx}
\usepackage{booktabs}
\usepackage{multirow}
\usepackage{caption}
\usepackage{flushend}

\def\BibTeX{{\rm B\kern-.05em{\sc i\kern-.025em b}\kern-.08em
    T\kern-.1667em\lower.7ex\hbox{E}\kern-.125emX}}

\begin{document}
\newcolumntype{P}[1]{>{\centering\arraybackslash}p{#1}}
\newcolumntype{M}[1]{>{\centering\arraybackslash}m{#1}}
\setlength{\textfloatsep}{10pt plus 1.0pt minus 3.0pt}
\setlength{\dbltextfloatsep}{10pt plus 1.0pt minus 3.0pt}
\setlength{\floatsep}{10pt plus 1.0pt minus 3.0pt}
\setlength{\dblfloatsep}{10pt plus 1.0pt minus 3.0pt}
\setlength{\intextsep}{10pt plus 1.0pt minus 3.0pt}

\title{\LARGE \bf STARNet: Sensor Trustworthiness and Anomaly Recognition via Approximated Likelihood Regret for Robust Edge Autonomy}
\author{Nastaran Darabi$^*$, Sina Tayebati$^*$, Sureshkumar S., Sathya Ravi, Theja Tulabandhula, and Amit R. Trivedi
\thanks{$^*$ Both authors contributed equally to this work. Authors are with the University of Illinois Chicago (UIC), Chicago, IL, Email: {\tt\small amitrt@uic.edu}}
}
\maketitle

\begin{abstract} 
Complex sensors such as LiDAR, RADAR, and event cameras have proliferated in autonomous robotics to enhance perception and understanding of the environment. Meanwhile, these sensors are also vulnerable to diverse failure mechanisms that can intricately interact with their operation environment. In parallel, the limited availability of training data on complex sensors also affects the reliability of their deep learning-based prediction flow, where their prediction models can fail to generalize to environments not adequately captured in the training set. To address these reliability concerns, this paper introduces STARNet, a Sensor Trustworthiness and Anomaly Recognition Network designed to detect untrustworthy sensor streams that may arise from sensor malfunctions and/or challenging environments. We specifically benchmark STARNet on LiDAR and camera data. STARNet employs the concept of approximated likelihood regret, a gradient-free framework tailored for low-complexity hardware, especially those with only fixed-point precision capabilities. Through extensive simulations, we demonstrate the efficacy of STARNet in detecting untrustworthy sensor streams in unimodal and multimodal settings. In particular, the network shows superior performance in addressing internal sensor failures, such as cross-sensor interference and crosstalk. In diverse test scenarios involving adverse weather and sensor malfunctions, we show that STARNet enhances prediction accuracy by approximately 10\% by filtering out untrustworthy sensor streams. STARNet is publicly available at \textcolor{magenta}{\emph{\url{https://github.com/sinatayebati/STARNet}}}.
\end{abstract}


\section{Introduction}
The demand for improved perception and a deeper understanding of the environment in autonomous robotics has led to an increased reliance on complex sensors. These sensors, such as those capable of capturing data beyond the visible spectrum, provide robots with a more detailed view of their surroundings. For instance, LiDAR sensors offer precise depth perception and spatial resolution, enabling detailed 3D mapping, which is valuable for object detection, mapping, and localization in autonomous navigation. LiDAR is also effective in various lighting conditions, including nighttime and overcast situations, where cameras struggle \cite{arikumar2022real}. Similarly, RADAR performs well in adverse weather conditions, reliably detecting objects in fog, rain, or snow and accurately measuring velocities. Consequently, there is a growing interest in utilizing combinations of these complex sensors to enhance object recognition, material differentiation, environmental monitoring, and related applications for \textbf{\textit{robust autonomy}} in diverse scenarios.

Meanwhile, acquiring sufficient training data for advanced sensors is challenging due to several factors. For instance, with the evolution of LiDAR technology, new sensor designs, such as optical phased array \cite{poulton2019long}, yield data with distinct characteristics. Keeping training data up-to-date with these advancements is intricate. Similarly, reflections from stationary objects like buildings, trees, or parked vehicles can influence RADAR sensor readings, making target differentiation in cluttered environments a significant labeling challenge. Additionally, active sensors require more diverse training data due to the unique influences of various environments and lighting conditions on return signals. These factors complicate developing exhaustive training datasets that cover all potential scenarios a robot might face in its operational life.

\begin{figure}[t!]
    \centering
    \includegraphics[width=\linewidth]{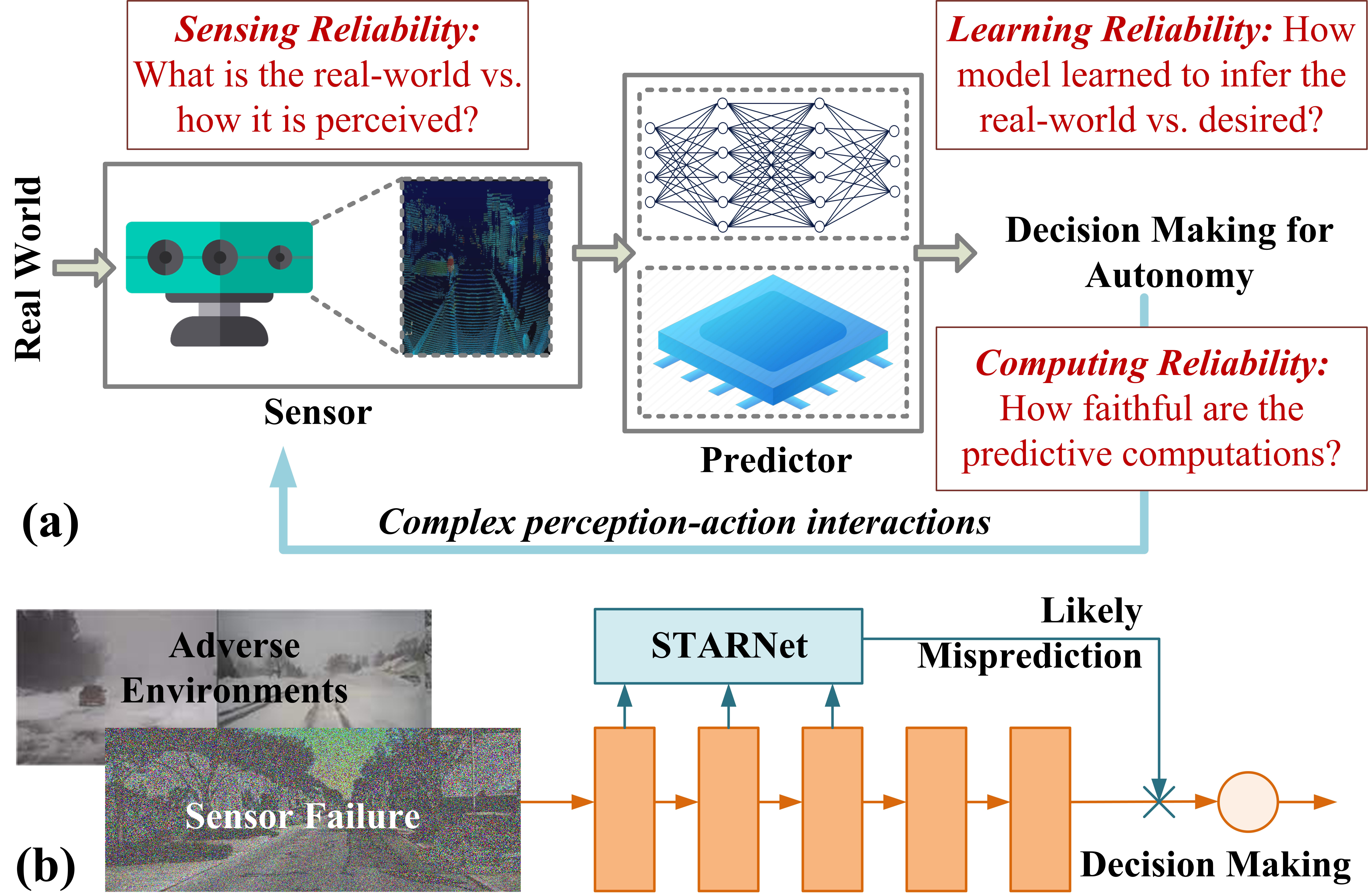}
    \caption{\textbf{Motivation for STARNet:} \textbf{(a)} Predictions for edge autonomy can be impacted by a lack of reliability in sensing, learning, and computing. Various failures may impact sensors, such as a beam missing in LiDAR. Hard-to-generalize environments impact learning reliability. Limited computing resources at the edge impact the reliability of computations such as those demanding high-precision operations. \textbf{(b)} In this work, we present STARNet, which feeds upon intermediate representations of the sensor streams, a function of sensor reliability and environment, to detect when the prediction failures are likely. STARNet is designed to operate under low precision, so edge computing constraints less impact it.}
    \label{fig:overview}
\end{figure}

Additionally, complex sensors are also susceptible to various failures. The demand for compact, energy-efficient sensors necessitates innovative fabrication techniques. As sensor complexity increases, the manufacturing process must achieve higher precision, which also raises the risk of internal component failures. For example, on-chip integration of edge-emitting lasers with nanometer node transistor technologies can result in complex failure mechanisms due to thermal overstress \cite{zhao2019impact}, electrostatic discharge \cite{yan2022reliability}, fabrication variabilities \cite{kalasapati2022robustness}, aging \cite{tan2019long}, \textit{etc.} Likewise, in emerging multi-pixel LiDAR and RADAR sensors \cite{diehm2018mitigation}, interference from nearby pixels can result in cross-talk, thus corrupting their sensor readings.

With the growing integration of complex sensors and advanced deep learning in next-generation robotics, ensuring the continuous reliability of sensor data streams becomes crucial. To address the challenges of robust autonomy, we introduce \textbf{STARNet}: a \textbf{S}ensor \textbf{T}rustworthiness and \textbf{A}nomaly \textbf{R}ecognition \textbf{Net}work. STARNet detects untrustworthy sensor features, which may arise from sensor malfunctions or difficult-to-generalize environments, to alert downstream decision-making processes of potential inaccuracies.

Within STARNet, we employ the concept of \textit{approximated likelihood regret} which is tailored for low-complexity hardware, especially those with only fixed-point precision capabilities, thus ensuring minimal impact from edge computing constraints. STARNet utilizes the generative capabilities of variational auto-encoders (VAE) to learn the distribution of trustworthy sensor data streams, which jointly depend on sensor reliability and the generalizability of the environment based on the training set. The likelihood of a sensor stream is computed using pre-trained VAE $L_\text{VAE}$ and then again by optimizing the VAE against the input sample,i.e., $L_\text{OPT}$ using the proposed gradient-free processing. The difference in likelihoods, $L_\text{OPT}-L_\text{VAE}$, termed \textit{likelihood regret}, effectively differentiates between trustworthy and untrustworthy streams: trustworthy streams display lower likelihood regret as they align closely with the learned distribution. In contrast, untrustworthy streams exhibit a pronounced regret. By continuously monitoring and filtering out untrustworthy sensor streams, we demonstrate that bSTARNet improves the prediction accuracy by $\sim$10\% across various test cases on adverse weather and sensor malfunctioning. 

\section{Background}

\subsection{Sensor Failures}
In autonomous navigation systems, sensor reliability directly influences the efficacy of deep learning-based decision-making. While complex sensors are becoming prevalent in autonomous robotics, they complicate reliability verification due to their intricate failure mechanisms. For instance, LiDAR sensors can experience ghost readings from multi-path interference, reduced accuracy during adverse weather like fog, or even false negatives due to the absorption of signals by raindrops \cite{heinzler2019weather}. Infrared sensors might be impacted by ambient temperature fluctuations or direct exposure to bright sources, leading to noisy or even incorrect readings \cite{gandhi2007pedestrian}. The emerging event-based cameras are impacted by rapid scene changes, causing a temporal information overload \cite{gallego2020event}. Modern solid-state RADARs can suffer from false reflections or clutter from non-moving objects \cite{sole2004solid}. Acoustic sensors, used for subterranean or underwater navigation, can face signal attenuation or reflection complications due to environmental inconsistencies \cite{li2022experience}. MEMS-based inertial sensors can drift over time or get affected by sudden temperature shifts \cite{kok2017using}. While precise modeling of these sensor faults is challenging, especially due to their complex interaction with the environment and autonomy task, even minor inaccuracies in the input data can lead to disproportionately large errors in the output due to the non-linearity of deep learning models.     

\subsection{Out-of-Distribution Detection (OOD)}
In STARNet, we utilize out-of-distribution (OOD) detection techniques to identify sensor streams that significantly deviate from the training data distribution. Various OOD detection methods exist. For instance, WarningNet \cite{lee2020warningnet} detects sensor anomalies due to alterations in image input patterns via deep learning, while the PAAD network \cite{ji2022proactive} proactively signals robots about anomalies in unpredictable environments. Generative models, such as Variational Autoencoders (VAEs) \cite{masuda2021toward} and Generative Adversarial Networks (GANs) \cite{bergmann2023anomaly}, are employed for OOD detection, using VAEs' reconstruction errors and GANs' discriminator scores as OOD indicators. Recently, there has been a surge of self-supervised learning methods \cite{hendrycks2019benchmarking, afham2022crosspoint} for OOD detection as well. 

However, many of these established methods, including Softmax probability scores \cite{hendrycks2016baseline} and feature representations from the network's last layer \cite{lee2018simple}, often fail to detect seemingly straightforward OOD scenarios, such as discriminating the distribution of CIFAR-10 dataset images from MNIST images. A key drawback of these methods is the potential overlap in the feature space of in-distribution and OOD data. If an OOD sample is proximate to the decision boundary in this space, it may be incorrectly identified as an in-distribution sample. Unlike earlier approaches, the likelihood regret (LR) measure \cite{xiao2020likelihood} has shown a much-improved ability to detect OOD data by training a distribution learner (e.g., VAE) on incoming sample to ascertain the discrepancy between the predicted likelihood from the primary model and that from a model retrained on this new sample. The LR values for OOD samples are typically substantially different than those for in-distribution samples, aiding discrimination. However, the computational cost of LR--owing to the need for retraining on every sample--constrains its use in real-time edge devices. We address this using gradient-free computation techniques, discussed next.

\begin{figure*}[t!]
    \centering
    \includegraphics[width=\linewidth,trim={0 17pt 0 0},clip]{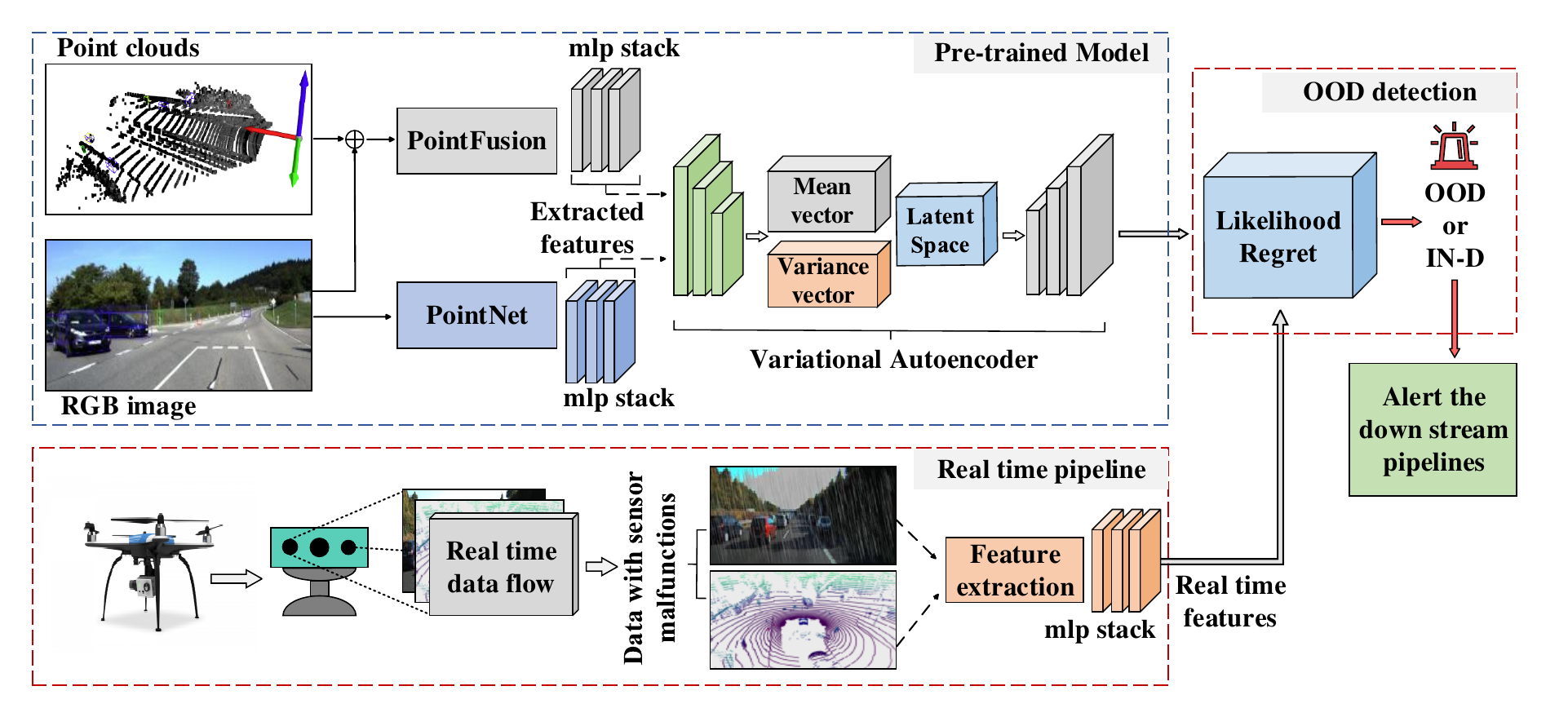}
    \caption{\textbf{STARNet on Multimodal Data Streams:} We demonstrate STARNet using multimodal sensor streams from cameras and LiDAR point clouds. However, STARNet's framework is versatile and can be adapted to other multimodal scenarios. In the current configuration, STARNet feeds upon sensor feature representations generated for the primary task, specifically from PointNet and PointFusion. A VAE ingests these features to learn their typical distribution. During inference, STARNet calculates the likelihood regret through gradient-free optimization to identify the discrepancy between the sensed feature and learned feature distribution to alert the primary prediction mechanism of potential inaccuracies when discrepancies are excessive.}
    \label{fig: Network}
\end{figure*}

\subsection{Gradient-Free Optimization}
In STARNet, we use gradient-free optimization suitable for fixed-point precision hardware. Traditional gradient-based methods rely on floating-point precision hardware to accurately represent a broad spectrum of gradient values. This increases energy and footprint due to extra circuitry for the mantissa, exponent processing, and error correction. On the other hand, fixed-point precision hardware, which uses a simpler arithmetic logic unit, is more common in edge devices because deep learning inference often works well with fixed-point computation. Therefore, by tailoring STARNet's optimization problems to be gradient-free, we present a more resource-efficient and widely applicable framework.

Among the prominent gradient-free optimization methods, ZO-SGD \cite{ghadimi2013stochastic} and ZO-SCD \cite{lian2016comprehensive} stand out for optimizing unconstrained stochastic optimization with convergence rates of \(O(\sqrt{d}/\sqrt{T})\), where \(T\) represents the iteration count. However, their convergence efficiency is hampered by the variable dimension \(d\). The ZO stochastic mirror descent method (ZO-SMD) was introduced to mitigate this, establishing a tighter bound with dimension-dependent factors \cite{phillips2017dust}. Additionally, variance reduction techniques have bolstered ZO-SGD and ZO-SCD, leading to stochastic variance-reduced algorithms that boast enhanced convergence rates and iteration complexities \cite{liu2018zeroth}. Recent ZO optimization algorithms, such as ZO proximal SGD (ZO-ProxSGD) \cite{ghadimi2016mini}, ZO via conditional gradient (ZO-CG) \cite{balasubramanian2018zeroth}, and online alternating direction method for multipliers (ZO-ADMM) \cite{gao2018information}, emphasize gradient-free constrained optimization. Another significant method in this domain is Simultaneous Perturbation Stochastic Approximation (SPSA), which efficiently estimates gradients by perturbing all dimensions simultaneously, allowing for scalability \cite{bhatnagar2023generalized}. 

\section{Framework of STARNet}
Fig. \ref{fig: Network} presents the overview of STARNet. STARNet operates on the intermediate sensor stream representation generated for downstream tasks, such as object detection. These representations are fed to a VAE to learn their distribution from the training set. Notably, the generated distributions are a function of \textit{both} the sensor reliability and the operating environment. By leveraging the same representations for the downstream task, STARNet presents minimal overheads. The reliability of an incoming sensor stream is verified by computing its likelihood regret (LR) using the proposed gradient approximation methods to determine if the sensor features emanate from the learned distribution or are out-of-distribution, in which case, the downstream controller is alerted for possible misprediction.

We characterize STARNet in two settings: LiDAR-only processing and LiDAR+Camera processing. For the LiDAR-only case, the PointNet architecture \cite{qi2017pointnet} is employed. Instead of relying on data representations like volumetric grids or multiple 2D projections, PointNet processes raw point coordinates directly, ensuring invariance under permutations of the input set. The network structure in PointNet employs a combination of shared multi-layer perceptrons (MLPs) and max-pooling layers to capture local features of individual points and global contextual information of the entire point set. For LiDAR+camera processing, PointFusion \cite{xu2018pointfusion} is employed for feature extraction and fusion. By aligning and integrating these heterogeneous data types, PointFusion exploits the spatial depth information from point clouds and the rich texture and color information from images. Even though we present our results on these two specific test cases, the framework of STARNet can be generalized to other extractors. 

\begin{figure}[t!]
    \centering
    \includegraphics[width=0.49\linewidth]{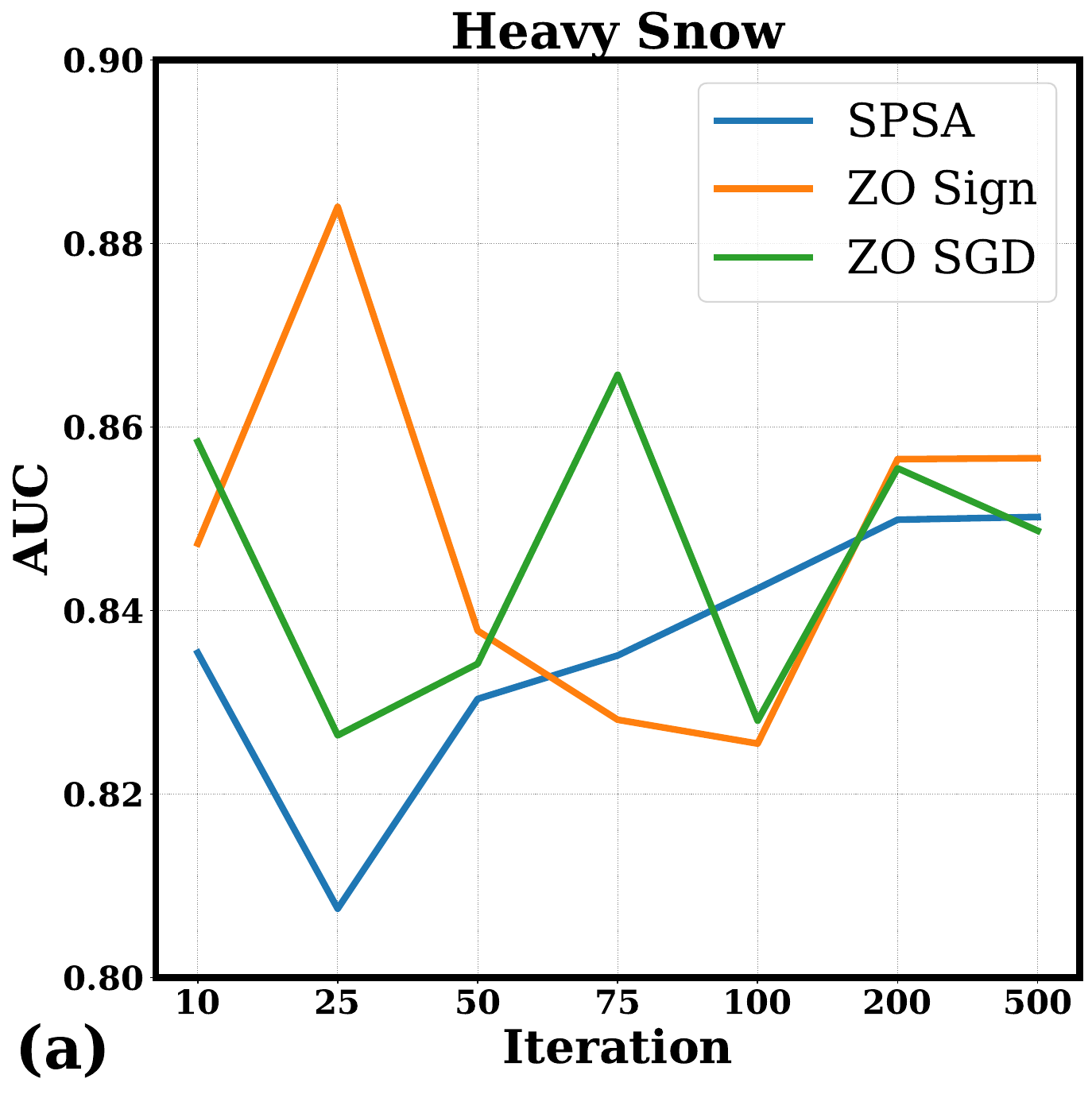}
    \includegraphics[width=0.49\linewidth]{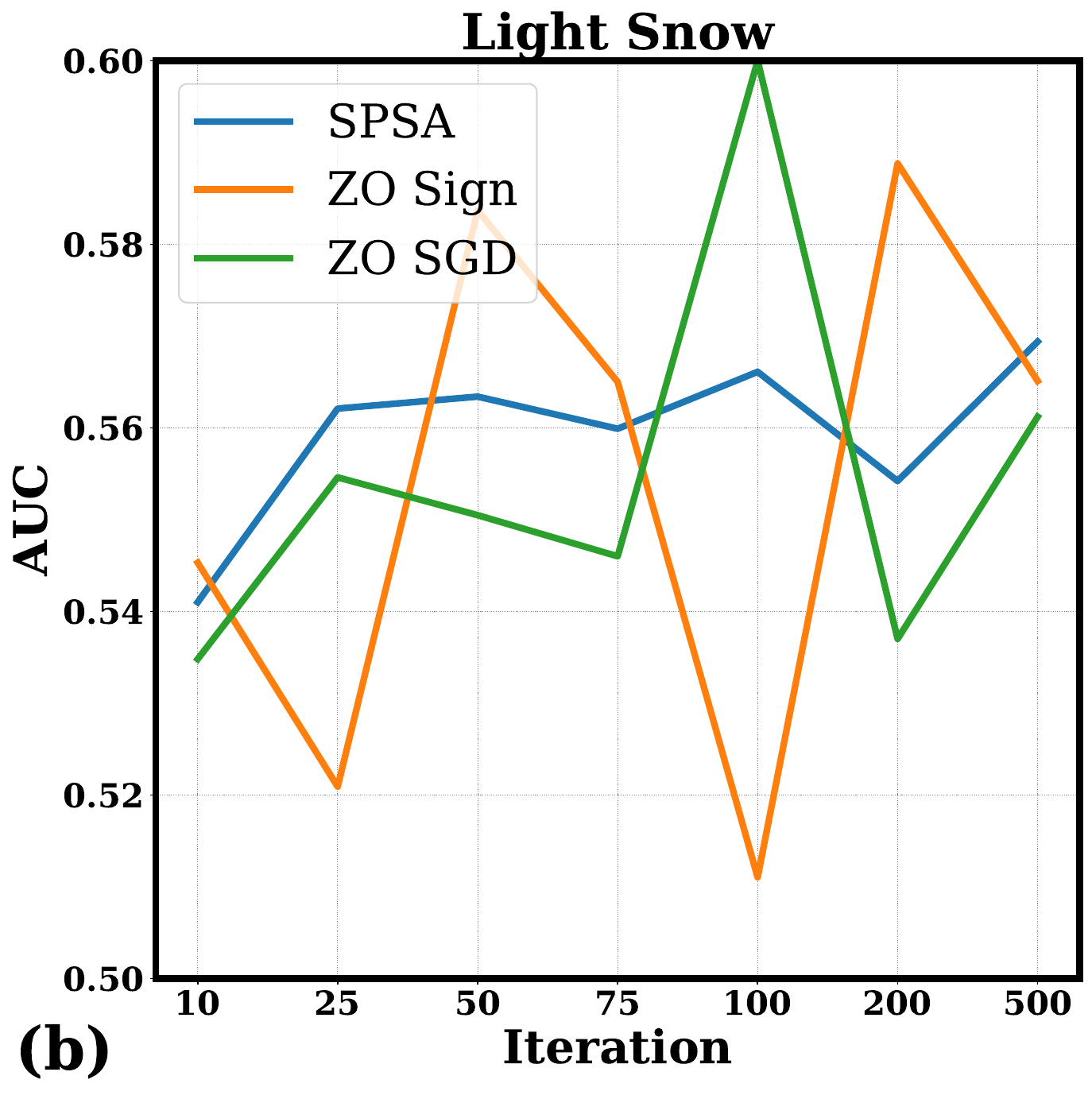}    
    \caption {\textbf{Gradient-free Optimization:} A comparative evaluation of three gradient-free optimization methods—Zero Order Sign, Zero Order Stochastic Gradient Descent, and Simultaneous Perturbation Stochastic Approximation (SPSA)—under two conditions: \textbf{(a)} heavy snow and \textbf{(b)} light snow. The graph shows noticeable fluctuations in the performance of Zero Order Sign and Zero Order Stochastic Gradient Descent as the number of iterations increases. At the same time, SPSA is highlighted for its enhanced robustness and stability.}
    \label{fig: ZO}
\end{figure}

\begin{figure}[t!]
    \centering
    \includegraphics[width=\linewidth]{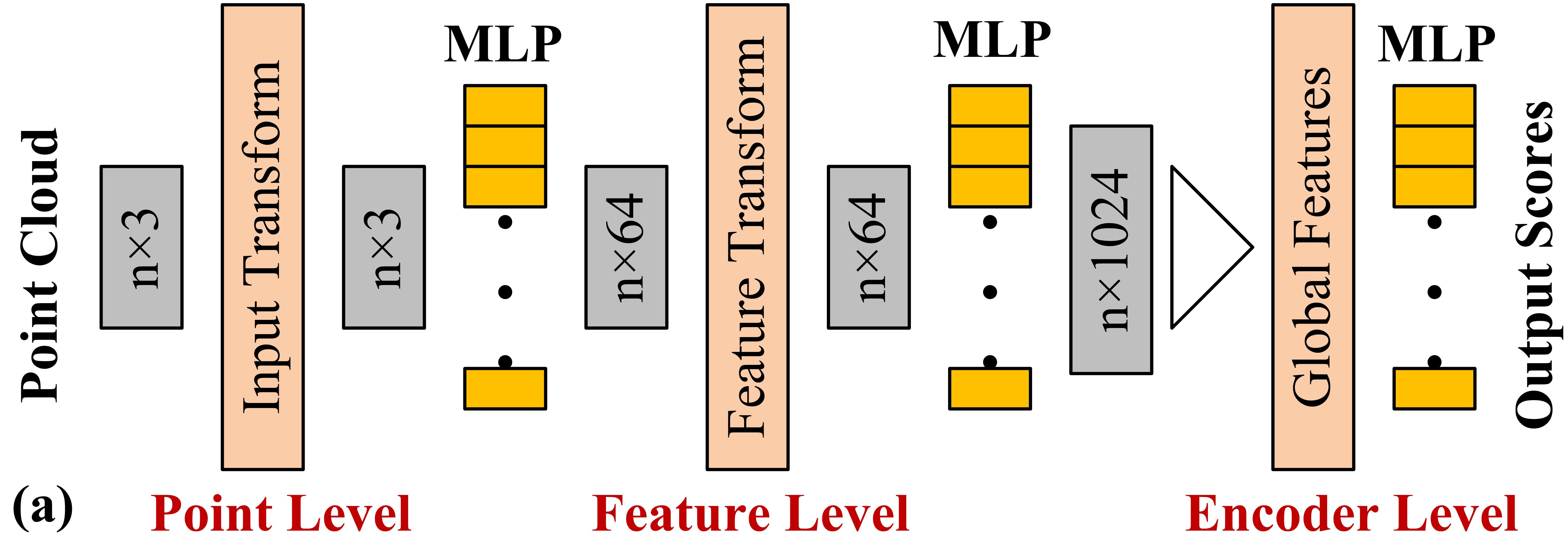}
    \includegraphics[width=\linewidth]{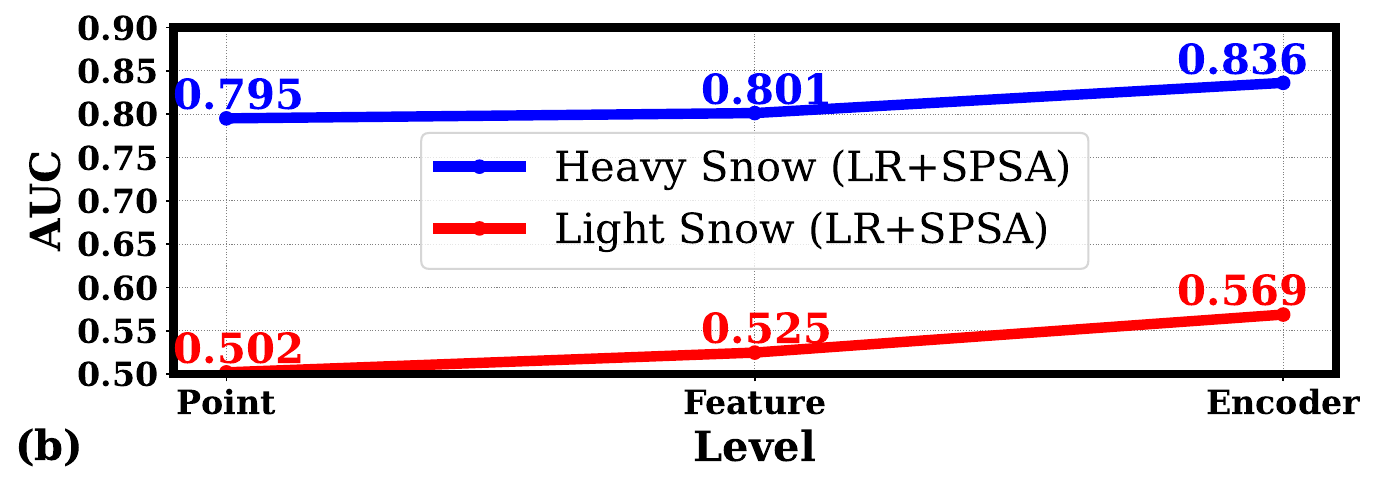}
    \caption{\textbf{Feature Extraction Analysis:} \textbf{(a)} High-level architecture of the PointNet version used for feature extraction, detailing its three main components: point level, feature level, and encoder level. \textbf{(b)} Comparative analysis of AUC values derived from features extracted at each level, highlighting the superior performance of encoder features under heavy and light snow conditions. }
    \label{fig: Feature}
\end{figure}

\begin{figure*}[t!]
    \centering
    \includegraphics[width=\linewidth]{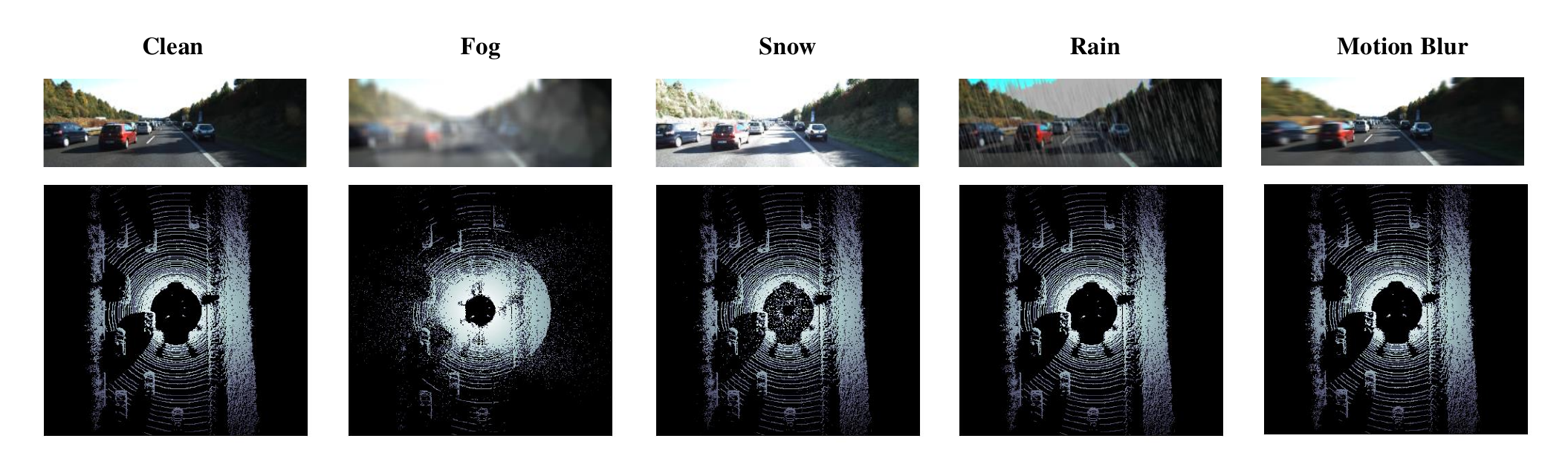}
    \caption {\textbf{Visualization of Corruption sets:} Our multimodal test-set contains the clean KITTI dataset and four different failure cases on both camera images and bird-view LiDAR point clouds.}
    \label{fig: corruption}
\end{figure*}

\subsection{Unsupervised Learning of Sensor Feature Distribution}
We utilize a variational autoencoder (VAE) to learn the distribution of extracted sensor features. This VAE models the observed variable \( x \) using a latent random variable \( z \). The generative model is represented as:
\begin{equation}
p(x) = \int p(x|z)p(z)dz
\end{equation}
For a computationally efficient training of VAE, we focus on maximizing the evidence lower bound (ELBO), given by:
\begin{equation}
\log(p_\theta(x)) \geq \mathbb{E}_{q_\phi(z|x)} \left[ \log p_\theta(x|z) \right] - D_{KL} \left[ q_\phi(z|x) || p(z) \right]
\end{equation}
Here, \(q_{\phi}(z|x) \) is the variational approximation to the true posterior \(p_{\theta}(z|x) \), with global parameters \(\phi\) (encoder) and \(\theta\) (decoder). The Kullback-Leibler divergence, \( D_{KL} \), measures the difference between the prior and the approximating distribution. The main goal is to train the VAE to maximize the ELBO across training data, ensuring the VAE captures the data's inherent distribution and offers a useful representation in the latent space. 

\subsection{Likelihood Regret under Gradient Approximation}
LR captures the discrepancy between an input's likelihood from its optimized posterior distribution and the likelihood predicted by the VAE, defined as $\text{LR} = L_\text{OPT} - L_\text{VAE}$, where $L_\text{OPT}$ is the empirical dataset (or sample) optimum and $L_\text{VAE}$ is the optimum corresponding to the input example. In-distribution samples should have lower LR values than out-of-distribution samples. We explored gradient approximation techniques for the computations of LR, differentiating it from the original method in \cite{xiao2020likelihood}. We specifically investigated several gradient approximation methods, including zero-order optimization with stochastic gradient descent (ZO\_SGD) \cite{zhao2020towards, liu2020primer}, zero-order optimization with sign \cite{cheng2019sign, liu2020primer}, and simultaneous perturbation stochastic approximation (SPSA) \cite{bhatnagar2023generalized}. 

Fig. \ref{fig: ZO} shows the comparison among methods using AUC (Area Under the ROC, i.e., Receiver Operating Characteristic curve) as the evaluation metric, a conventional metric for assessing the performance of OOD classification. Compared to other methods, SPSA shows improved performance on a more challenging light snow case and also consistently benefits with more iterations of gradient-free model updates, making it our preferred choice. Additionally, the method is also lightweight because it only requires evaluating the system at two perturbed points, regardless of the dimensionality of the problem. Also, considering the feature extraction flow of PointNet in Fig. \ref{fig: Feature}(a), we also analyzed the optimal abstraction level that results in the best performance. PointNet comprises three components: point, feature, and encoder levels. Upon extracting features from each component, our analysis showed that encoder features yielded the highest AUC values, as presented in Fig. \ref{fig: Feature}(b). These findings are utilized in our simulation results discussed next.

\section{Simulation Results}

\subsection{Benchmark Testsets}
To characterize the efficacy of STARNet, we developed benchmark testing datasets featuring eight distinct corruption types, each presented in two different severity levels (Heavy and moderate). These datasets are divided into the following categories: (i) \textit{Natural Corruptions:} Environmental factors like fog, rain, and snow, (ii) \textit{External Disruptions:} Issues such as motion blur and beam missing (absence of laser beams), and (iii) \textit{Internal Sensor Failures:} Sensor-specific problems including crosstalk, incomplete echo responses, and cross-sensor interference. Though our approach utilizes LiDAR-Camera fusion, all corruption scenarios affected both data types. However, the internal sensor failures pertained solely to the LiDAR sensor. We evaluated our method using the KITTI-C dataset, which provides corruption sets for LiDAR and camera data \cite{kong2023robo3d}. Figure \ref{fig: corruption} showcases dataset samples.

\begin{figure*}[t!]
    \centering
    \includegraphics[width=\linewidth,trim={0 0 0 20pt},clip]{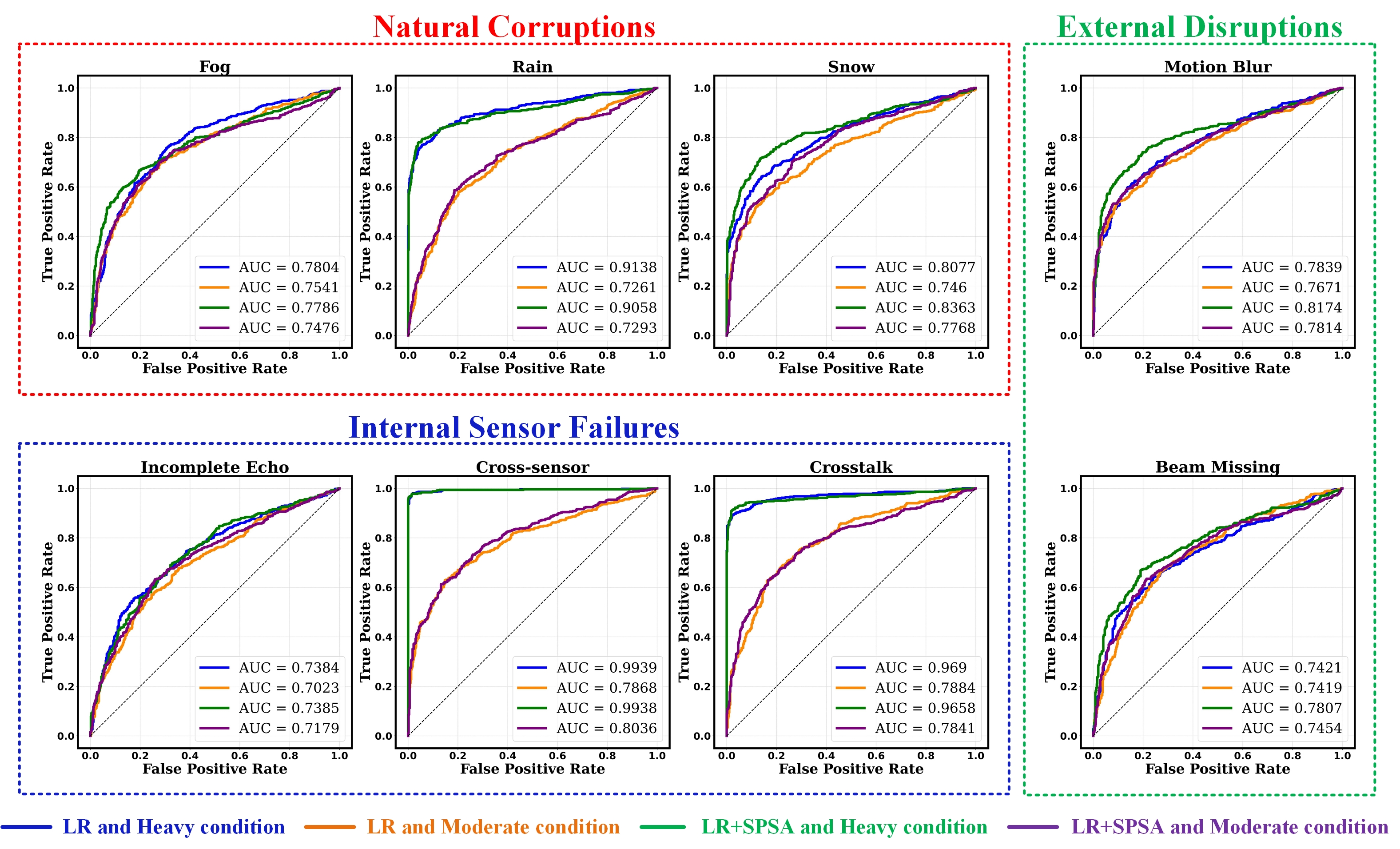}
    \caption{\textbf{Unimodal results:} Performance evaluation of STARNet on LiDAR point clouds for eight different failure cases. The figure highlights the superior performance of our method in addressing internal sensor failures, specifically cross-sensor interference and crosstalk. Notably, the AUC values for heavy crosstalk and cross-sensor interference are 0.9658 and 0.9938, respectively.}
    \label{fig:Lidar_features}
\end{figure*}

We evaluated STARNet's efficacy in unimodal and multimodal settings considering these corruptions. In the unimodal setting, only the LiDAR point cloud was processed. The point cloud was combined with its corresponding image in the multimodal setting. In the following, we discuss our simulation results for both cases.

\subsection{Results for LiDAR-only Processing}
We examined eight failure scenarios for LiDAR point clouds, each presented in heavy and moderate intensities. As shown in Fig. \ref{fig:Lidar_features}, our method performs well, especially when dealing with internal sensor issues like cross-sensor interference and crosstalk. Specifically, our method achieved an AUC value of 0.9658 in cases with intense crosstalk. For cross-sensor interference, the AUC value was 0.9938.

Table I provides a detailed comparison of our method against various Out-of-Distribution (OOD) detection techniques, including the original Likelihood Regret\cite{xiao2020likelihood}, Log-Likelihood, Likelihood Ratio\cite{ren2019likelihood}, and the Input Complexity method\cite{serra2019input}. The results in the table show that our approach, which uses an approximate gradient-free likelihood regret with SPSA, is almost as effective as the original method that relies on gradient descent. Additionally, these methods perform better than other techniques, such as the likelihood ratio and input complexity.

\begin{table*}
\small
    \setlength{\tabcolsep}{6pt}
    \renewcommand{\arraystretch}{1.2}
    \centering 
    \captionsetup{font=small}
    \caption{\textbf{Comparison Results for Unimodal case (Metric: AUC)}}
    \begin{tabularx}{\linewidth}{c||cccccc}
        \hline
              & \textbf{Ours (LR+SPSA)} & \textbf{Likelihood Regret\cite{xiao2020likelihood}} & \textbf{Likelihood} & \textbf{Likelihood Ratio\cite{ren2019likelihood}} & \textbf{Input Complexity\cite{serra2019input}} \\
        \hline
        Fog & 0.7786 & 0.7804 & 0.4294 & 0.7612 & 0.4009\\
        Snow & 0.8363 & 0.8077 & 0.3613 & 0.7884 &  0.3587\\
        Rain & 0.9058 & 0.9138 & 0.4248 & 0.9216 & 0.3910 \\
        Motion Blur & 0.8174 & 0.7839 & 0.3972 & 0.7407 & 0.3914\\
        Beam Missing & 0.7807 & 0.7421 & 0.4551 & 0.7531 & 0.4407\\
        Incomplete Echo & 0.7385 & 0.7384 & 0.3839 & 0.7079 &  0.3635\\
        Cross-sensor & 0.9938 & 0.9939 & 0.4180 & 0.8743 & 0.3734\\
        Crosstalk & 0.9658 & 0.969 &  0.3087 & 0.9235 & 0.2780\\        
        \hline
    \end{tabularx}
    \begin{flushleft}
        The pre-trained model is only on the clean KITTI point cloud. Thus, the results are obtained under unsupervised learning.         
    \end{flushleft}
    % \hline
\end{table*}

\subsection{Results for Camera + LiDAR Processing}
\begin{figure*}[t!]
    \centering
    \includegraphics[width=\linewidth,trim={0 0 0 15pt},clip]{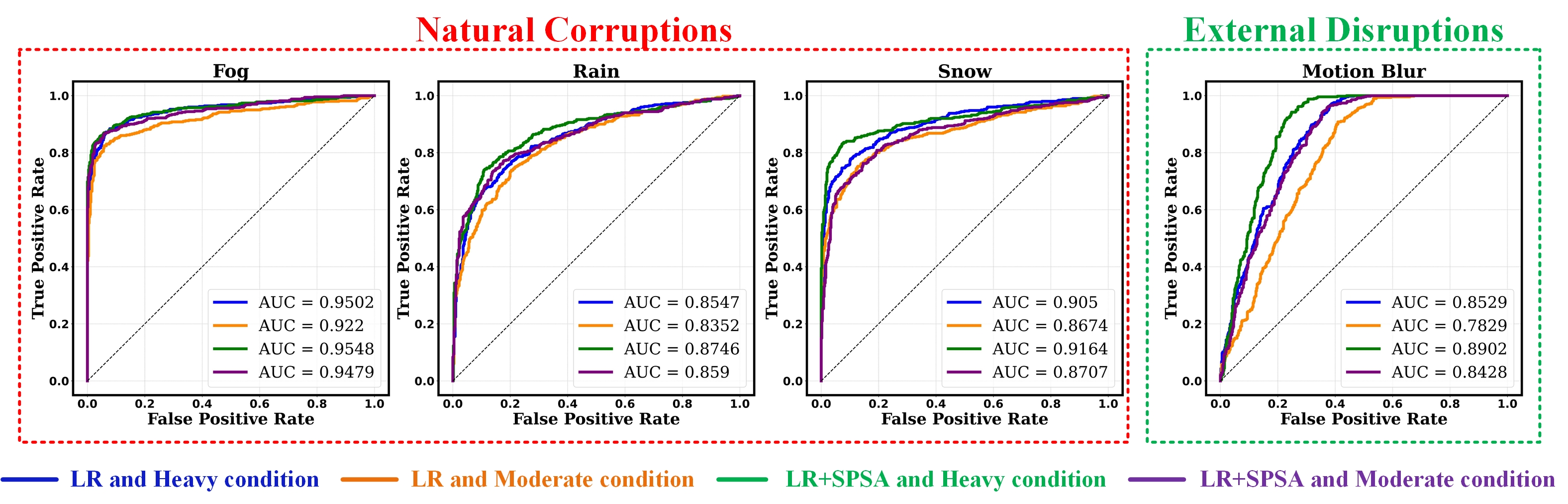}    
    \caption{\textbf{Multimodal results:} Performance evaluation of STARNet on LiDAR point clouds and images for four different failure cases. }
    \label{fig:Multimodal}
\end{figure*}

In the context of multimodal scenarios, we undertook an analysis of four distinct failure situations concerning the fusion of LiDAR point clouds and images. Two levels of intensity characterized each of these scenarios. As illustrated in Fig. \ref{fig:Multimodal}, our method's performance closely aligns with the original LR's. A comprehensive summary of our findings is presented in Table II. The table incorporates accuracy values based on the R2 score. This metric underscores the notion that, while there might be minor discrepancies in OOD detection, the overall task can still be executed with high precision. Our results indicate that integrating images and LiDAR data enhances the AUC values. For instance, in the presence of snow, the AUC increases by 0.0801 under heavy snow. Similarly, AUC increases by 0.0939 under moderate snow conditions.

\begin{table}
\small
    \setlength{\tabcolsep}{6pt}
    \renewcommand{\arraystretch}{1.2}
    \centering 
    \captionsetup{font=small}
    \caption{\textbf{Fusion Results (Metric: AUC)}}
    \begin{tabularx}{\linewidth}{c||ccc||c}
        \hline
              & \textbf{LR+SPSA} & \textbf{LR} & \textbf{Log-LL} & \textbf{ ACC}\\
        \hline
        H Fog & 0.9548 & 0.9502 & 0.2860 & 0.9663 \\
        M Fog & 0.9479 & 0.922 & 0.2846 & 0.9667 \\
        %L Fog & 0.7421 & 0.7114 & 0.3078 & 0.9663 \\
        \hline
        H Snow & 0.9164 & 0.905 & 0.3283 & 0.9756\\
        M Snow & 0.8707 & 0.8674 & 0.3974 & 0.97\\
        %L Snow & 0.869 & 0.8344 & 0.3805 & 0.9723 \\
        \hline
        H Rain & 0.8745 & 0.8547 & 0.351 & 0.9798\\
        M Rain & 0.8590 & 0.8352 & 0.4042 & 0.9799\\
        %L Rain & 0.8468 & 0.8305 & 0.374 & 0.9831\\
        \hline
        H Motion Blur & 0.8902 & 0.8529  & 0.6058  & 0.9750\\
        M Motion Blur & 0.8428 & 0.7829  & 0.6238  & 0.9763\\
        %L Motion Blur & 0.7628 & 0.7036  & 0.6351  & 0.9729\\
        \hline
    \end{tabularx}
    \begin{flushleft}
        Accuracy (ACC) is reported as the R2 score. \\
        The pre-trained model is on the Clean KITTI point cloud.          
    \end{flushleft}
    % \hline
\end{table}

\begin{figure}[t!]
    \centering
    \includegraphics[width=\linewidth]{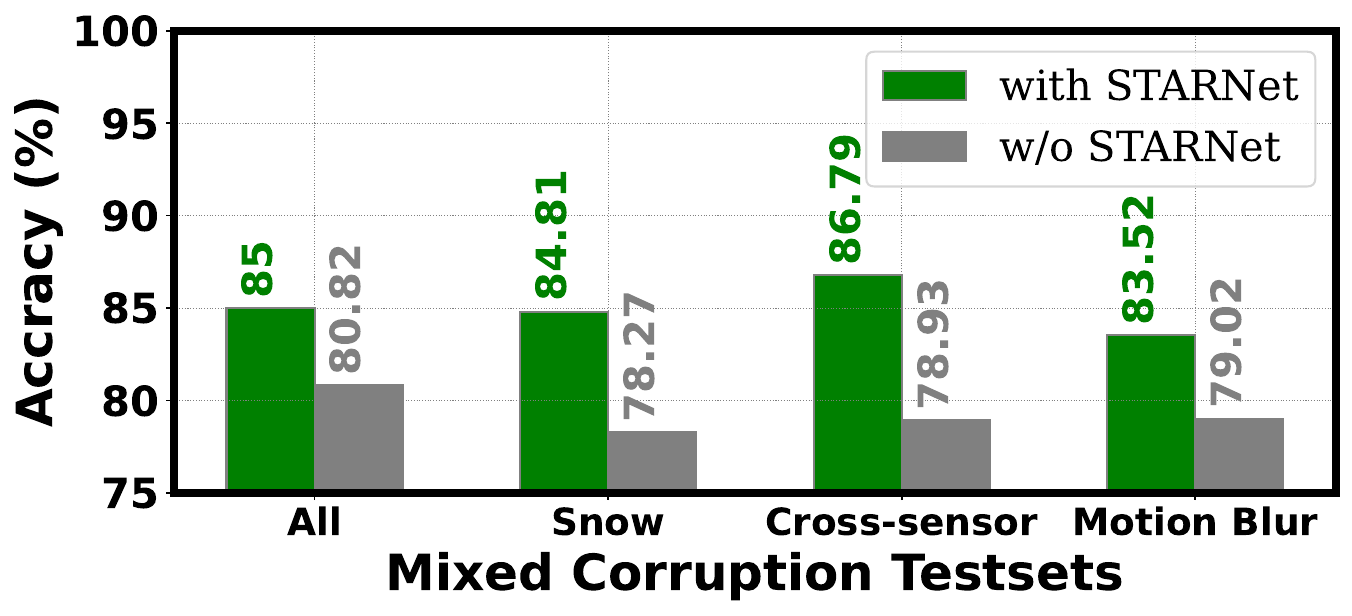}    
    \caption{\textbf{Object Detection Accuracy for KITTI dataset:} We employed a VAE to process LiDAR point clouds and corresponding labels of various objects, including cars, pedestrians, and cyclists. The VAE then outputs bounding boxes for these objects. The network was tested on challenging conditions, including heavy, medium, and low snow and other corruption cases.}
    \label{fig:ACC}
\end{figure}

\section{Conclusion}
Although deep learning models can be enhanced for robustness by augmenting training data with noisy sets and diverse operating environments, solely relying on augmentation is insufficient since it is hard to anticipate and gather enough training data to capture the complexity and diversity of environments encountered in practical scenarios. Comparatively, STARNet enables risk-conscious decision-making by continuously monitoring the trustworthiness of incoming sensor streams. In Fig. 8, we summarize the net significance of STARNet in improving robust decision-making. Utilizing a VAE trained solely on clean data for object detection, we observed its susceptibility to several environmental and sensor-based corruptions, including snow, cross-sensor noise, and motion blur. These disturbances result in a decline from its original prediction accuracy of approximately 87.32\%. With STARNet's integration to detect and mitigate untrustworthy sensor data, the prediction accuracy in these challenging scenarios improved by $\sim$10\%, effectively reverting to the level observed with clean data. Moreover, STARNet's reliance on computationally efficient gradient approximation techniques ensures its viability for continuous monitoring, even on resource-constrained edge devices.

% \ifCLASSOPTIONcaptionsoff
%     \newpage
% \fi
\bibliographystyle{IEEEtran}
\bibliography{main}

\end{document}